\documentclass{article}
\usepackage{arxiv}
\usepackage[utf8]{inputenc} % allow utf-8 input
\usepackage[T1]{fontenc}    % use 8-bit T1 fonts
\usepackage{hyperref}       % hyperlinks
\usepackage{url}            % simple URL typesetting
\usepackage{booktabs}       % professional-quality tables
\usepackage{amsfonts}       % blackboard math symbols
\usepackage{nicefrac}       % compact symbols for 1/2, etc.
\usepackage{microtype}      % microtypography
\usepackage{lipsum}
\usepackage{fancyhdr}       % header
\usepackage{graphicx}       % graphics
\graphicspath{{media/}}     % organize your images and other figures under media/ folder
\usepackage[sort&compress,numbers]{natbib}

\hypersetup{
    colorlinks=true,
}

\title{CEPHA29: Automatic Cephalometric Landmark Detection Challenge 2023}

\author{
    {Muhammad Anwaar Khalid}\textsuperscript{1} \\
    \And
    {Kanwal Zulfiqar}\textsuperscript{2} \\
    \And
    {Ulfat Bashir}\textsuperscript{2} \\ 
    \And
    {Areeba Shaheen}\textsuperscript{2} \\
    \And
    {Rida Iqbal}\textsuperscript{2} \\
    \And
    {Zarnab Rizwan}\textsuperscript{2} \\
    \And
    {Ghina Rizwan}\textsuperscript{2} \\
    \And
    {Muhammad Moazam Fraz}\textsuperscript{1} \\
    \And
    \textnormal{
        \textsuperscript{1}{National University of Sciences and Technology (NUST), Islamabad, Pakistan}} \\
    \textnormal{
        \textsuperscript{2}{Riphah International University, Islamabad, Pakistan}} \\
    {\tt cepha29.challenge@gmail.com}
}

\begin{document}
    \maketitle

    \begin{abstract}
    Quantitative cephalometry is the most widely used clinical and research tool in modern orthodontics, enabling the quantification and classification of anatomical abnormalities through localization of cephalometric landmarks. However, the traditional practice of marking these landmarks manually is a very tedious and error prone. Endeavors have constantly been made to develop automated cephalometric landmark detection systems but they are inadequate for orthodontic applications. The fundamental reason for this is that the amount of publicly available datasets as well as the images provided for training in these datasets are insufficient for an AI model to perform well. To facilitate the development of robust AI solutions for morphometric analysis, we organize the CEPHA29 Automatic Cephalometric Landmark Detection Challenge in conjunction with IEEE International Symposium on Biomedical Imaging (ISBI 2023). In this context, we provide the most diverse and comprehensive cephalometric dataset to date, consisting of 1000 lateral cephalometric radiographs (LCRs) obtained from 7 different radiographic imaging devices with varying resolutions. We hope our challenge will inspire innovative research and foster breakthroughs in automatic cephalometric landmark identification, ushering in a new era of progress in the field.
    \end{abstract}

    % keywords can be removed
    
    \section{Introduction}
    \label{sec:introdution}
    Cephalometric analysis of two-dimensional (2D) X-ray images, often referred to as cephalograms, is a fundamental investigation in modern orthodontics, orthognathic treatment and maxillofacial surgery \cite{proffit2006contemporary}. It provides essential information about the spatial relationships of teeth, jaws and cranial base and assists clinicians in diagnosing the craniofacial condition of a patient. Clinical applications of cephalometric analysis in surgical planning include assessment of upper and lower jaws \cite{nanda1994cephalometric, lake1981surgical, melnik1992cephalometric}, diagnosis and treatment of obstructive sleep apnea \cite{deberry1988cephalometric, guilleminault1988women, hochban1994morphology}, and assessment of soft facial tissue \cite{hajeer2004three, mcdonnell1977advancement}.

    In clinical settings, the landmarks are usually traced out manually by professional orthodontists, which is a time-consuming \cite{kamoen2001clinical} and unreliable approach to achieving reproducible results \cite{da2006reproducibility}. Moreover, several reports have also voiced concerns about the significant inter- and intra-observer variabilities among orthodontists \cite{durao2015cephalometric} due to their different training and experience backgrounds. Since pathology identification and subsequent treatment procedures are highly sensitive to precise estimation of landmark locations, poor manual cephalometric analysis might have serious repercussions. Therefore, it is highly desired to develop fully automated and reliable computerized frameworks that can detect cephalometric landmarks, perform required measurements, and assess anatomical abnormalities precisely and quickly.
    
    Throughout the recent decades, as technology has advanced in deep learning and computer vision, several research studies have been conducted on computer-aided landmark detection. Following this, Wang et al. organized the very first Automatic Cephalometric X-Ray Landmark Detection Challenge under IEEE International Symposium on Biomedical Imaging (ISBI) in both 2014 \cite{wang2014grand} and 2015 \cite{wang2015evaluation}, and summarized the performance of state-of-the-art landmark detection algorithms \cite{wang2016benchmark}. However, a comprehensive review of the existing literature revealed that the automatic success detection rate of anatomical landmarks has only increased from 71.47\% to 82.03\% within 2.0 mm, which is the clinically accepted precision range for a landmark prediction. This is primarily because of the limited availability of adequate medical imaging datasets for cephalometric analysis and high cost of annotations. Consequently, the AI models trained over currently available datasets are a bit sensitive to over-fitting and have shown poor performance on test data \cite{arik2017fully, lee2020automated, zeng2021cascaded}. Thus, developing a fully automated landmark detection framework capable of detecting all landmarks within the clinically accepted range is still an open challenge.
    
    To foster innovation and progress in automated quantitative cephalometry, we organize the Automatic Cephalometric Landmark Detection Challenge (CEPHA29)\footnote{CEPHA29: http://vision.seecs.edu.pk/cepha29/} in conjunction and with the help of IEEE International Symposium on Biomedical Imaging (ISBI 2023). To help participants in training their AI algorithms, we provide a state-of-the-art dataset \cite{khalid2023aariz} consisting of 1000 cephalometric X-ray images acquired from 7 different radiographic imaging devices. Each cephalogram in the dataset has been labelled by clinical experts with the 29 most commonly used cephalometric landmarks. In addition, the experts also marked the cervical vertebra maturation (CVM) Stage of the patient corresponding to each image. We are optimistic that this challenge will advance research and innovation in automated quantitative cephalometry, enabling the development of robust AI algorithms for precise anatomical landmark detection and CVM stage prediction.

    \section{Challenge Details}
    \label{sec:challenge-details}
    In the following section, we will provide a comprehensive overview of the CEPHA29 Automatic Cephalometric Landmark Detection Challenge 2023, including the technical details regarding the tasks, their corresponding descriptions, and specifications of the dataset that participants will be utilizing throughout the challenge.
        
        \subsection{Tasks}
        In this challenge, the participants are required to develop algorithms that can perform the following two tasks:
        
            \begin{enumerate}
                \item Automatic localization of 29 anatomical landmarks on the cephalogram in the clinically accepted precision range of 2.0 mm.
                \item Classification of the patient’s cephalogram in one of the six CVM stages. 
            \end{enumerate} 

\begin{figure*}[ht]
                \centering
                \includegraphics[width=0.95\textwidth]{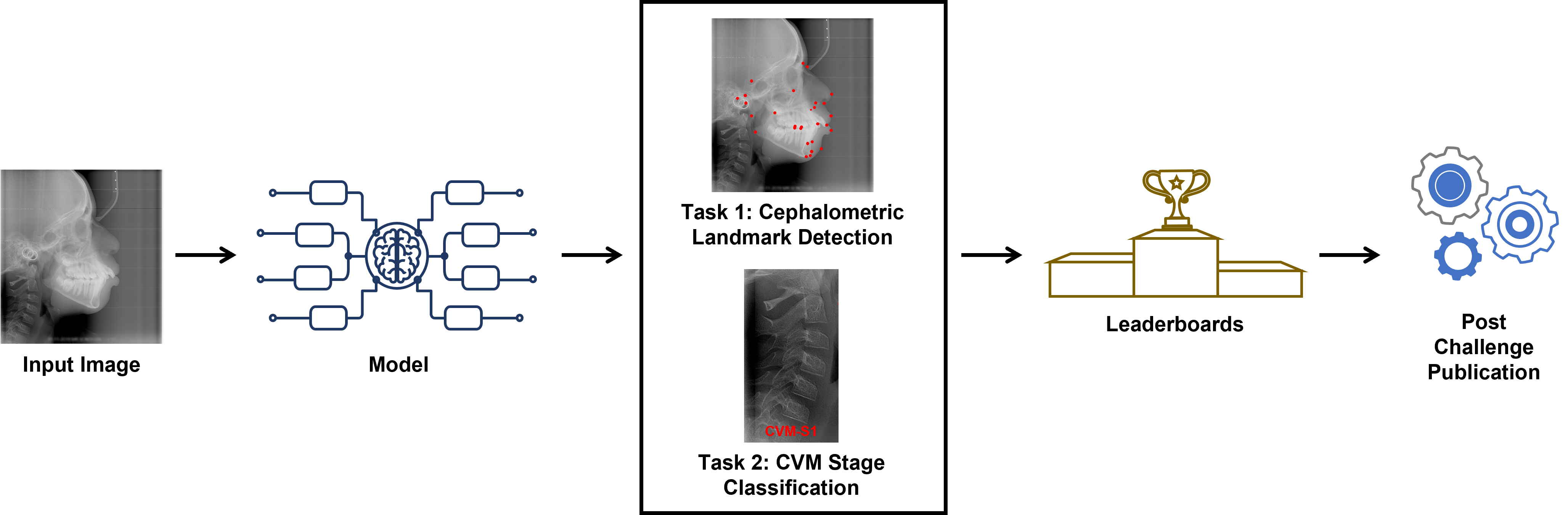}
                \caption{A high-level depiction of the task execution pipeline for the CEPHA29 Automatic Cephalometric Landmark Detection Challenge 2023, showcasing the overall flow of the challenge tasks and their execution.}
                \label{figure:overview-challenge-tasks}
            \end{figure*}

        In recent years, several research studies \cite{arik2017fully, chen2019cephalometric, zhong2019attention, qian2019cephanet, qian2019cephanet, oh2020deep, qian2020cephann, lee2020automated, zeng2021cascaded, kwon2021multistage, wang2021dcnn, he2021cephalometric} have emerged concerning the cephalometric landmark detection problem. As these approaches have shown promising results on the ISBI 2015 dataset \cite{wang2016benchmark}, we strongly recommend our participants to consider the artistic design and intuition of these techniques while developing their algorithms. It is important to note that the type of algorithm developed will determine its capability to perform the two tasks. For instance, a model designed for landmark identification can accomplish both tasks, whereas a model developed to predict the CVM stage cannot naturally locate anatomical anchor points. Therefore, we encourage out participants to review the latest proposed methods \cite{makaremi2019deep, seo2021comparison, atici2022fully, mohammad2022deep, makaremi2020determination} for determining the CVM stage from lateral cephalometric radiographs to ensure they develop robust algorithms that can accurately perform the required tasks. An overview of the tasks during the challenge is illustrated in Figure \ref{figure:overview-challenge-tasks}.
            
        \subsection{Data}
        % \subsubsection{\emph{\textquotesingle Aariz} Dataset}
        The challenge is based on a state-of-the-art dataset \cite{khalid2023aariz} consisting of 1000 cephalometric X-ray images from patients aged between 8 to 62 years. The X-ray images have been obtained from 7 different radiographic imaging devices with varying resolutions, which makes the dataset more diverse and challenging for the participants. Each cephalogram has been annotated with 29 cephalometric landmarks that are commonly used in clinical settings for a wide range of cephalometric measurements. Additionally, the CVM stage of each patient in the cephalogram has also been marked, providing valuable information for predicting the skeletal maturation of the patient.

        The dataset has undergone rigorous annotation by expert orthodontists. In the first phase, two junior orthodontists independently marked all the cephalograms, while in the second phase, two senior orthodontists collaboratively reviewed the markings and corrected them as necessary. The average of the markings from both the junior and senior orthodontists are provided separately, and the mean of these two averaged markings is considered as the ground truth for cephalometric landmarks.

        \subsubsection{Training Data}
        The Training set comprises of 700 lateral cephalometric X-ray images, each accompanied by detailed annotations. These annotations include:
        \begin{itemize}
            \item A JSON file containing the point coordinates of 29 cephalometric landmarks referring to the markings of junior orthodontists.
            \item Another JSON file containing the point coordinates of the same 29 cephalometric landmarks but referring to the markings of senior orthodontists.
            \item A JSON file holding the label for the CVM stage of the patient in that cephalogram.
        \end{itemize}
        To ensure that the training set is well-balanced and accurately reflects the diversity of clinical scenarios., the images from each of the 7 sources have been given equal representation. Moreover, a CSV file is provided with information about the X-ray imaging device used for each cephalogram, along with its resolution in terms of pixel size (mm). This information can be used by the participants to evaluate the performance of their algorithms.
            
        \subsubsection{Validation Data}
        The validation set is a critical component of the challenge, enabling participants to evaluate the performance of their algorithms on previously unseen data. It consists of 150 lateral cephalometric X-ray images without annotations, requiring participants to predict the position of 29 cephalometric landmarks and the CVM stage of each patient. Participants are required to submit their prediction results for this validation set in a specific format, and their performance will be evaluated based on the metrics discussed in Section \ref{sec:submission-evaluation}. The highest-performing teams will be selected to move on to the final round. This Evaluation Phase-1 provides an opportunity for participants to identify any potential issues with their algorithms and refine them before submitting their final results.
            
        \subsubsection{Testing Data}
        The Test set is comprised of 150 lateral cephalometric X-ray images that will be used to evaluate the performance of the participants' algorithms and determine the final rankings. The rankings will be based on the performance of the submitted algorithms on both the validation and Test set, as evaluated by the metrics discussed in Section 3. The Test set will not be provided to participants, and its annotations will be kept secret until the end of the challenge to ensure a fair evaluation of the algorithms. As this challenge uses code-based submissions, participants will be required to submit their containerized algorithms for final evaluation. All teams must submit their code to qualify for the final round, and those that do not will be disqualified. The submission process will be open for a specific period after the release of the validation set, and participants will be notified about the submission timeline in advance.

\begin{figure*}[t]
            \centering
            \includegraphics[width=\textwidth]{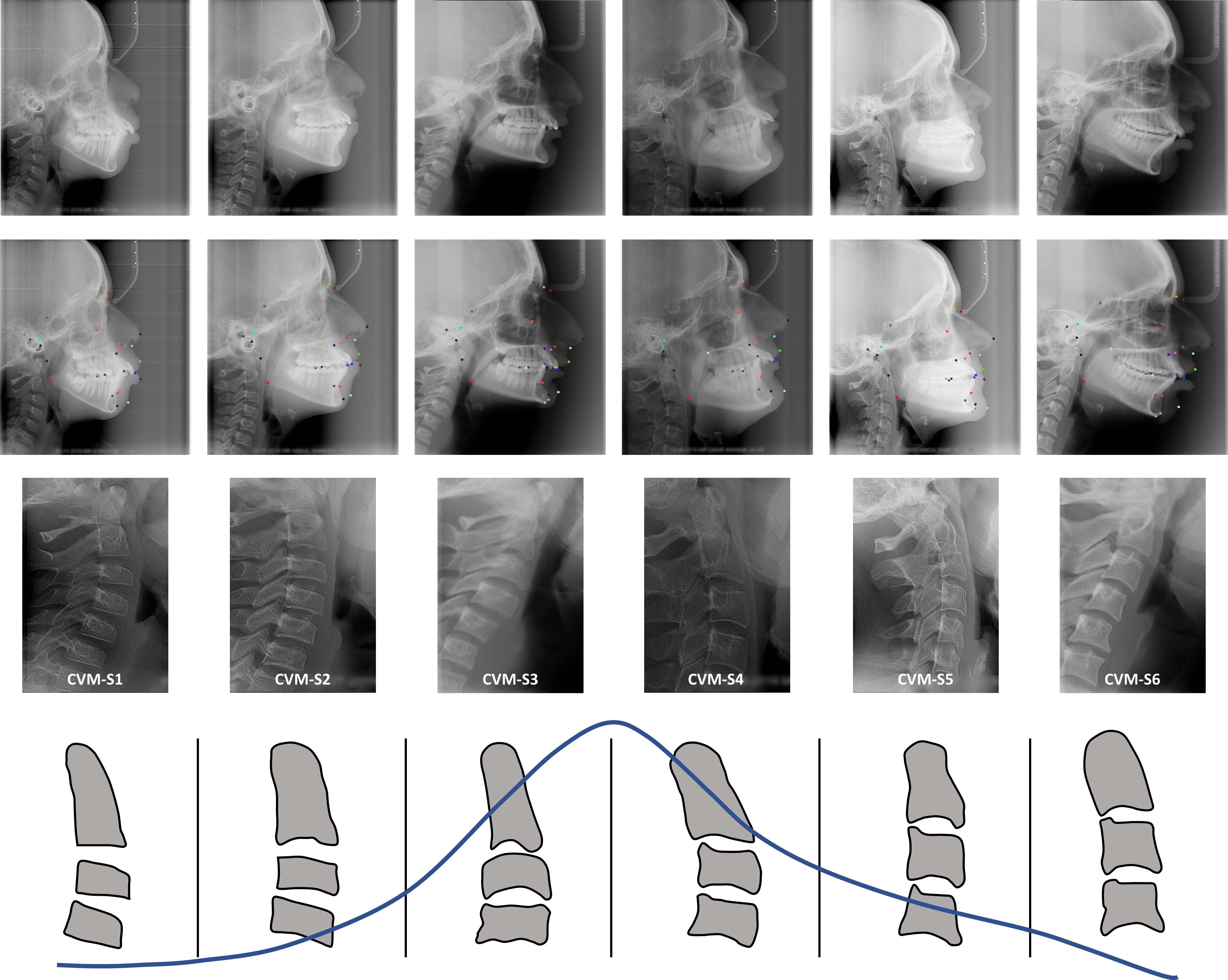}
            \caption{Examples of cephalograms from the dataset with corresponding ground truth annotations. (\textbf{top}) Cephalograms along with the respective ground truth for 29 cephalometric landmarks and the patient's CVM stage. (\textbf{bottom}) The radiological and morphological stages of CVM superimposed with the Björk growth curve \cite{ELHADDAOUI2014293}}.
            \label{dataset-example-images}
        \end{figure*}

        We have included some examples of cephalograms with their corresponding cephalometric landmarks and CVM stage in Figure \ref{dataset-example-images}. Participants are encouraged to refer to a detailed article \cite{khalid2023aariz} for more information on the dataset. To help participants get started with the challenge tasks, we have provided code on our GitHub page\footnote{https://github.com/manwaarkhd/CEPHA29} that can read the cephalograms along with their annotations and perform all the necessary operations.
            
    \section{Submission and Evaluation}
    \label{sec:submission-evaluation}
        In the following section, we will outline the evaluation metrics that will be used to assess the performance of the proposed algorithms for each task. These metrics will determine the final rankings of submissions on both leaderboards. To assist participants in evaluating their models, we have also made available evaluation code on our GitHub page that includes the specified metrics for thorough testing. Additionally, we will provide instructions on the format and content of the short report that participants are required to submit describing their proposed method in detail.
        
        \subsection{Evaluation Metrics}        
        \subsubsection{Cephalometric Landmark Detection:}
            The effectiveness of the proposed method for landmark detection will be evaluated based on two main statistical measures. The first evaluation criterion is the mean radial error, which measures the average distance between the predicted and ground truth landmarks. The radial error is calculated using the following formula:
            \begin{equation}
                R = \sqrt{(\Delta x)^{2} + (\Delta y)^{2}}
            \end{equation}
            where $\Delta x$ and $\Delta y$ are the absolute distances in the x and y directions, respectively, between the predicted and ground truth landmark positions. The Mean Radial Error (MRE) and its associated Standard Deviation (SD) are calculated as follows:
            \begin{equation}
                MRE = \frac{1}{N} \sum_{i=1}^{N} R_i
            \end{equation}
            \begin{equation}
                SD = \sqrt{\frac{\sum_{i=1}^{N}(R_i-MRE)^{2}}{N-1}}
            \end{equation}
            where $N$ is the total number of landmarks and $R_i$ is the radial error for the $i$-th landmark.
            The second evaluation criterion is the Success Detection Rate (SDR) with a precision range of 2.0 mm, 2.5 mm, 3.0 mm, and 4.0 mm. For each landmark, a reference location is marked by orthodontists as a single pixel. A detected landmark is considered successful if the absolute difference between the detected and ground truth landmark positions is within the clinically accepted range, and otherwise, it is considered a detection failure. The success detection rate $p_z$ with precision less than $z$ is calculated as follows:
            \begin{equation}
                p_z = \frac{\{\#i: \parallel L_d(i)-L_g(i) \parallel < z \}}{\# \Omega} \times 100
            \end{equation}
            where $L_d$ and $L_g$ represent the positions of the detected and ground truth landmarks, respectively, $z$ denotes the precision range, and $\# \Omega$ is the total number of detections made.
            
        \subsubsection{CVM Stage Classification:}
            Four widely-used statistical measures will be employed to evaluate the performance of the proposed method for CVM stage classification. The first metric is accuracy, which measures how close a given set of measurements are to their corresponding true values. It can be expressed as:
            \begin{equation}
                Accuracy = \frac{TP + TN}{TP + FP + TN + FN}
            \end{equation}
            where TP, FP, TN, and FN are the numbers of true positives, false positives, true negatives, and false negatives, respectively.
            
            The second metric is precision, which measures the proportion of correctly classified positive samples out of all samples classified as positive. It is computed as:
            \begin{equation}
                Precision = \frac{TP}{TP + FP}
            \end{equation}
            
            The third metric is recall, which is also known as sensitivity and measures the proportion of actual positive samples that are correctly classified as positive. It can be calculated as:
            \begin{equation}
                Recall = \frac{TP}{TP + FN}
            \end{equation}
            
            The fourth metric is the F1 score, which is the harmonic mean of precision and recall, and provides a balanced assessment of the classification model. It is given by:
            \begin{equation}
                F_1\;Score = 2 \times \frac{P \times R}{P + R}
            \end{equation}
            where P is precision and R is recall.
        
    \subsection{Manuscript Submission Format}
    Along with the code, participants are required to submit a short report in IEEE ISBI format, which should be no longer than 4 pages. The report should include the following sections:
    \begin{enumerate}
        \item \textbf{Introduction and Related Work}: The introduction should provide an overview of the problem statement and a brief description of the state-of-the-art methods that have been proposed for the problem. End this section with a clear statement of the research questions addressed in the submission.
        \item \textbf{Methodology}: This section should provide a clear and concise description of the proposed method. The description should include a list of techniques that were evaluated or tried and the details of the method that was selected and submitted to the challenge.
        \item \textbf{Results}: This section should provide a detailed explanation of the experimental evaluation of the selected method on the validation set. The results should be presented in a clear and concise manner, and if applicable, the section should also include ablation studies.
        \item \textbf{Discussion}: This section should provide a discussion of the outcomes and the implications of the results. It should also highlight the strengths and weaknesses of the proposed method and how it compares to the state-of-the-art methods.
    \end{enumerate}

    We encourage all participants to create multi-institutional teams and make their methods publicly available to support transparency and reproducibility, and to be leveraged by the clinical community. Based on the submitted methods and results, we may encourage some teams to potentially collaborate to achieve better results by combining methods.
    
    \section{Organization}
        In this section, we provide a tentative timeline for the challenge, along with details on the code of conduct and the procedure for post-challenge publication. The timeline includes key dates for registration, data release, submission deadlines, and announcement of results. The code of conduct outlines the ethical guidelines that all participants must adhere to during the challenge. Finally, the procedure for post-challenge publication provides details on how the results of the challenge will be disseminated to the wider research community.
        
        \subsection{Timeline}
\begin{figure*}[ht]
            \centering
            \includegraphics[width=\textwidth]{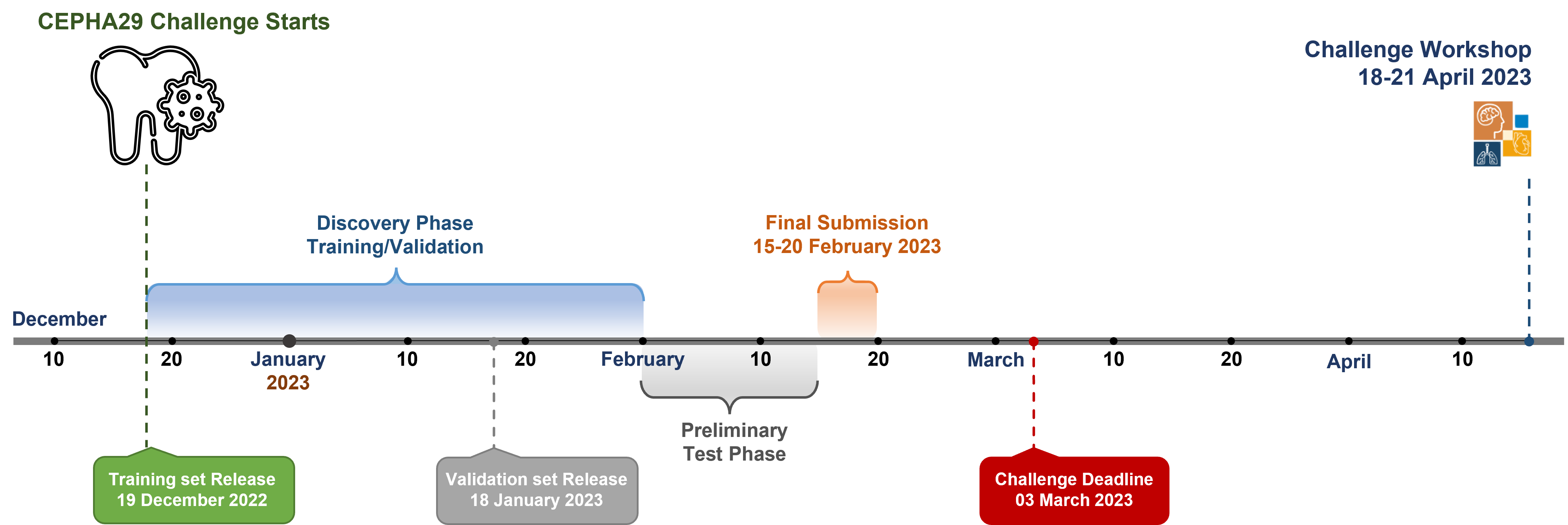}
            \caption{A chronological depiction of the planned challenge proceedings.}
            \label{fig:challenge-proceedings-timeline}
        \end{figure*}

        Table \ref{tab:challenge-schedule} presents a tentative schedule of the CEPHA29 Automatic Cephalometric Landmark Detection Challenge, which outlines the various phases of the challenge along with their corresponding key dates. We understand the importance of adhering to a timeline and will make every effort to ensure that the schedule is met. In case of any change in the schedule, the participants will be notified promptly through our challenge website. To provide a visual representation of the challenge proceedings, Fig \ref{fig:challenge-proceedings-timeline} depicts the timeline for each stage of the challenge.

        \begin{table}[htbp]
            \centering
            \caption{Timeline of the challenge phases and key dates.}
            \begin{tabular}{lc}
                \toprule
                \textbf{Description} & \textbf{Date} \\
                \midrule
                Registration opens for CEPHA29 challenge                        & \textbf{17 December 2022} \\ 
                Release of training data and evaluation code                    & \textbf{19 December 2022} \\ 
                Release of validation data                                      & \textbf{18 January 2023}  \\ 
                Submission for preliminary evaluation phase (Phase-1) starts    & \textbf{23 January 2023}  \\ 
                Registration deadline for team formation                        & \textbf{31 January 2023}  \\ 
                Submission deadline of results for evaluation (Phase-1)         & \textbf{10 February 2023} \\ 
                Final submission of containerized algorithms starts             & \textbf{16 February 2023} \\ 
                Submission deadline of containerized algorithms (Phase-2)       & \textbf{25 February 2023} \\
                Submission of IEEE ISBI double column paper                     & \textbf{01 March 2023}    \\
                Decision for paper acceptance                                   & \textbf{05 March 2023}    \\
                \bottomrule
            \end{tabular}
            \label{tab:challenge-schedule}
        \end{table}

        \subsection{Participation Rules and Code of Conduct}
        We have listed a set of guidelines below to ensure a fair and professional conduct throughout the challenge. Please read these rules carefully before registering for the challenge:
        \begin{itemize}
            \item All challenge participants must follow the scientific code of conduct and maintain professionalism at all times. Any misuse or inappropriate use of challenge-related material, on or off the website, is strictly prohibited. In case of any misconduct, please report it to the organizers immediately.
            \item Participants must agree to use the provided data following the scope of the Data Use and Confidentiality Agreement (DUA). Creating multiple accounts to download the data and submit the results is not permitted.
            \item External data cannot be used for training the network. However, pre-trained models with ImageNet \cite{deng2009imagenet} or MS COCO \cite{lin2014microsoft} datasets can be used. Pre-trained models with any other dataset are not allowed. Similarly, user-generated labels cannot be used.
            \item Final submissions must be in the form of Docker containers and must adhere to the template provided by the organizers. It is the user’s responsibility to ensure that their algorithm’s Docker container is in good working order during the preliminary test period before submitting it for the final test set.
            \item Only fully automated methods will be accepted for the final submission as submissions will be graded solely based on their performance on the final test set, and following the evaluation criteria explained in Section \ref{sec:submission-evaluation}.
            \item A technical report explaining the data split, pre-processing, model architectural design, post-processing, and any other necessary details related to the proposed solution must be written and submitted.
            \item Participants must agree to include the results of their proposed methods in a comprehensive review article. The results and their comparisons will be used in the preparation of a joint publication, with the challenge organizers(s) leading the write-up process.
            \item Users can participate in teams, and teams may have several submissions as long as they differ significantly from the methodology described in technical papers. Only one submission is allowed for each user/team for the final test set.
            \item The challenge organizing committee reserves the right to remove/eliminate teams at any point if they find any dispute, disrespectful emails, provocations, unnecessary email flood, incorrect leaderboard results, etc.
            \item The decisions of the organizing committee will be considered final and binding on all matters about this challenge.
        \end{itemize}
        
        \subsection{Post Challenge Publication}
        After the conclusion of the challenge, the top-performing authors will be invited to contribute to a joint publication, which will showcase their work in detail and provide an opportunity to share insights gained from the challenge. The publication will feature the goals, outcomes, and lessons learned from the challenge, and will be submitted to a top-tier journal. The selection of the journal will be based on the challenge outcomes and feedback from domain experts, ensuring that the publication is relevant and impactful in the field of automatic cephalometric landmark detection. This joint publication will serve as a valuable resource for researchers and practitioners interested in this area, and will help to advance the state-of-the-art in automatic cephalometric landmark detection.

        \section{Conclusion}
        The CEPHA29 Automatic Cephalometric Landmark Detection Challenge aims to accelerate the development of automatic cephalometric landmark detection methods and provide a benchmark for future research. We hope that this challenge will attract a diverse set of participants and foster collaboration and innovation in the field. We encourage interested researchers and practitioners to participate and contribute to this effort. The challenge timeline, participation rules, and evaluation criteria have been outlined in this paper, and we look forward to receiving submissions and sharing the results with the community. We believe that this challenge will lead to the development of more accurate and efficient cephalometric landmark detection methods, which will ultimately improve the diagnosis and treatment of craniofacial disorders.

% Bibliography
\bibliographystyle{unsrt}  
\bibliography{references}  

\end{document}